# Improving Incremental Nonlinear Dynamic Inversion Robustness Using Robust Control in Aerial Robotics

Mohamad Hachem[1], Clément Roos[2], Thierry Miquel[1], Murat Bronz[1]
[1]*ENAC-Lab, Dynamic Systems, Ecole Nationale de l'Aviation Civile*, Toulouse, France
[2]*Département Traitement de l'Information et Systèmes, ONERA, Université de Toulouse*, Toulouse, France

***Abstract*—Improving robustness to uncertainty and rejection of external disturbances represents a significant challenge in aerial robotics. Nonlinear controllers based on Incremental Nonlinear Dynamic Inversion (INDI), known for their ability in estimating disturbances through measured-filtered data, have been notably used in such applications. Typically, these controllers comprise two cascaded loops: an inner loop employing nonlinear dynamic inversion and an outer loop generating the virtual control inputs via linear controllers. In this paper, a novel methodology is introduced, that combines the advantages of INDI with the robustness of linear structured $\mathcal{H}_\infty$ controllers. A full cascaded architecture is proposed to control the dynamics of a multirotor drone, covering both stabilization and guidance. In particular, low-order $\mathcal{H}_\infty$ controllers are designed for the outer loop by properly structuring the problem and solving it through non-smooth optimization. A comparative analysis is conducted between an existing INDI/PD approach and the proposed INDI/$\mathcal{H}_\infty$ strategy, showing a notable enhancement in the rejection of external disturbances. It is carried out first using MATLAB simulations involving a nonlinear model of a Parrot Bebop quadcopter drone, and then experimentally using a customized quadcopter built by the ENAC team. The results show an improvement of more than 50% in the rejection of disturbances such as gusts.**

## I. Introduction

Aerial robotics has been successfully integrated into diverse fields, ranging from inspection and photography to package delivery as well as search & rescue operations. Each application presents its unique challenges, originating from the hardware framework or the software algorithms employed. Flying in a confined indoor environment can for example generate disturbances caused by the backwash from the drone's propellers. Flying with a cable-suspended payload or attached manipulators can disrupt the system, leading to external perturbations and oscillations. In general, taking into consideration disturbance attenuation objectives is essential. A controller that ensures precise positioning while attenuating external perturbations is therefore crucial for the success and safety of a mission.

Various types of controllers have been explored and evaluated to enhance the system's robustness and overall performance, starting with linear ones. The authors of [1] implement Proportional-Integral-Derivative (PID) controllers to perform aggressive maneuvers, which are designed in parallel with a dynamically feasible trajectory parameterized by a goal state. A comparison between Linear Quadratic Regulator (LQR) and PID controllers is presented in [2]. It is shown that for a quadcopter, an LQR controller has better robustness than a classical PID controller, while a PID tuned using an LQR loop improves the robustness and tracking speed of the system. However, linear controllers can usually guarantee robustness and performance only when operating sufficiently close to the linearization point, since multirotor dynamics exhibit considerable nonlinearities. Robust nonlinear controllers can therefore represent a viable alternative to linear ones. The authors of [3] adopt a quasi Linear Parameter Varying (quasi-LPV) approach to drone modeling and introduce an $LPV/\mathcal{H}_\infty$ controller to ensure the drone's robustness against unmodeled uncertainties and disturbances. In [4], a robust controller that limits control actuator inputs is coupled with an observer estimating external disturbances as well as unknown nonlinearities inherent to the real system's operations. Sliding mode controllers, with diverse formulations, are deployed in [5]–[7] to reject disturbances and improve tracking performances. As a final example, feedback linearization techniques are used to control fully-actuated and under-actuated multirotor systems in [8] and [9] respectively.

In particular, Incremental Nonlinear Dynamic Inversion (INDI) has attracted attention in the domain of flight control systems [10], [11]. It involves two control loops: the inner one performs the system's dynamic inversion, while the outer one is responsible for generating virtual control inputs using linear controllers. It has been successfully implemented and evaluated in aerial robotics in [12], [13], demonstrating its effectiveness in enhancing system robustness and mitigating external disturbances. An approach for fine-tuning the PID gains of the outer loop is proposed in [14], showing connections with time delay control. Various improvements to the INDI control loop are also introduced by [15]–[17] to enhance robustness against modeling uncertainties and manage time delays between different control loops. However, these modifications focus solely on the inner loop, without investigating the effect of the accompanying linear controller. But if the INDI controller is already capable of handling output disturbances, its robustness could be further enhanced by integrating into the outer loop a controller synthesized using a robust approach. This strategy would indeed ensure that the closed-loop system remains robust in both linear and nonlinear regimes. Among linear control techniques, $\mathcal{H}_\infty$ control is known for its ability to shape the system's closed-loop and to compute the controller's gains so as to satisfy certain predefined requirements, in particular the rejection of external perturbations. Structured $\mathcal{H}_\infty$ controllers have already been successfully designed in a previous work to control fully actuated hexacopters [18]. The integration of a $\mathcal{H}_\infty$ controller in addition to the existing INDI architecture could therefore play a crucial role in drastically attenuating the impact of external perturbations. Such a combined INDI/$\mathcal{H}_\infty$ control strategy has already been proposed for for guided projectiles applications [19], while nonlinear dynamic inversion (NDI) combined with $\mathcal{H}_\infty$ control was applied to aircraft applications by [20], [21]. But to the best of our knowledge, INDI/$\mathcal{H}_\infty$ has neither been applied to control multirotor systems, nor compared with existing methods.

In this context, a comprehensive methodology is proposed in this note to further enhance the robustness of the INDI approach by focusing on generating robust virtual control inputs. After linearizing the system with the INDI controller, the closed-loop sensitivity functions are shaped with the main objective of mitigating the effects of external perturbations, while preserving good performance levels, avoiding reaching actuator limits and filtering measurement noise. Instead of assuming that the disturbances are applied to the drone's output, as is usually done with INDI, they are applied to the system's input while designing the linear controller. It is then shown that a careful choice of the linear controller gains can significantly improve robustness. A structured $\mathcal{H}_\infty$ control problem is formulated and solved using non-smooth optimization techniques to obtain a low-order controller, which is scarcely more complex than the PID controllers classically proposed in the literature. The performances of this controller are compared with those of the full-order controller obtained by using the classical convex formulation of the $\mathcal{H}_\infty$ control problem, as well as the modal-based PD controller proposed in [12], [13]. The rest of this note is organized as follows. A brief introduction to INDI is provided in Section II-A, while the formulation of the considered $\mathcal{H}_\infty$ control problem is outlined in Section II-B. Section III then recalls the dynamical modeling of a typical quadcopter. The development and implementation of the proposed coupled INDI/$\mathcal{H}_\infty$ control architecture is discussed in Section IV-A for the stabilization loop and in Section IV-B for the guidance loop, including simulation results and comparisons with existing control strategies such as INDI/PD. An experimental validation of the entire architecture is finally reported in Section V, which highlights its ability to strongly mitigate the effect of gust-type perturbations.

## II. Theoretical background

The control architecture introduced in Section IV uses both INDI and structured $\mathcal{H}_\infty$ control. For the sake of clarity and completeness, the underlying theory is briefly outlined in this section.

### A. Incremental Nonlinear Dynamic Inversion

Incremental Nonlinear Dynamic Inversion (INDI) have been extensively presented in the literature by [22], and successfully applied to quadcopters by [12], [13]. Consider a nonlinear input-affine Multiple-Input Multiple-Output (MIMO) system with $n$ states $\boldsymbol{x}$ and $m$ inputs $\boldsymbol{u}$:

$$\dot{\boldsymbol{x}} = f(\boldsymbol{x}) + g(\boldsymbol{x})\boldsymbol{u} \tag{1}$$

A first-order Taylor expansion of the system dynamics is performed around the condition at the last sampling moment, marked in the sequel by the subscript 0:

$$\begin{aligned}\dot{\boldsymbol{x}} = \dot{\boldsymbol{x}}_0 + \left.\frac{\partial[f(\boldsymbol{x}) + g(\boldsymbol{x})\boldsymbol{u}_0]}{\partial \boldsymbol{x}}\right|_{\boldsymbol{x}=\boldsymbol{x}_0} (\boldsymbol{x} - \boldsymbol{x}_0) \\ + g(\boldsymbol{x}_0)\,(\boldsymbol{u} - \boldsymbol{u}_0)\end{aligned} \tag{2}$$

Since the controller usually runs at a high frequency in aerial robotics applications (greater than 500 Hz), it can be assumed that the change of states is negligible between two consecutive samples, i.e. that $\delta\boldsymbol{x} = \boldsymbol{x} - \boldsymbol{x}_0$ is almost zero. The resulting INDI control law is therefore defined as:

$$\boldsymbol{u} = \boldsymbol{u}_0 + g(\boldsymbol{x}_0)^{\dagger}(\boldsymbol{\nu} - \dot{\boldsymbol{x}}_0) \tag{3}$$

where † denotes the Moore Penrose pseudo-inverse. The effectiveness $g(\boldsymbol{x}_0)$ is usually estimated offline or identified online in practice, and $\boldsymbol{\nu}$ represents the desired dynamics of the state. It can be seen in equation (3) that the control input does not depend on the model information contained in $f(\boldsymbol{x})$. This makes the closed-loop system robust to model uncertainties.

### B. Structured $\mathcal{H}_\infty$ Control

$\mathcal{H}_\infty$ control aims to find a linear controller that stabilizes a dynamical system and minimizes the impact of exogenous inputs on closed-loop performance metrics. This problem can be written in the Linear Fractional Transformation (LFT) framework, as depicted in Fig. 1, where each considered metric is associated with a transfer function between an exogenous input $w_i$ and an exogenous output $z_j$. Solving the general $\mathcal{H}_\infty$ control problem then formally consists of calculating the controller $K(s)$ that stabilizes the generalized plant $P(s)$, usually composed of the system model and some weighting templates, and minimizes the performance index $\gamma$ under the following $\mathcal{L}_2$ induced norm constraint:

$$\|\boldsymbol{z(t)}\|_2 \leq \gamma \|\boldsymbol{w(t)}\|_2 \tag{4}$$

This is equivalent to minimizing the $\mathcal{H}_\infty$ norm of the transfer function $T_{\boldsymbol{w}\to\boldsymbol{z}}(s)$ between $\boldsymbol{w}$ and $\boldsymbol{z}$.

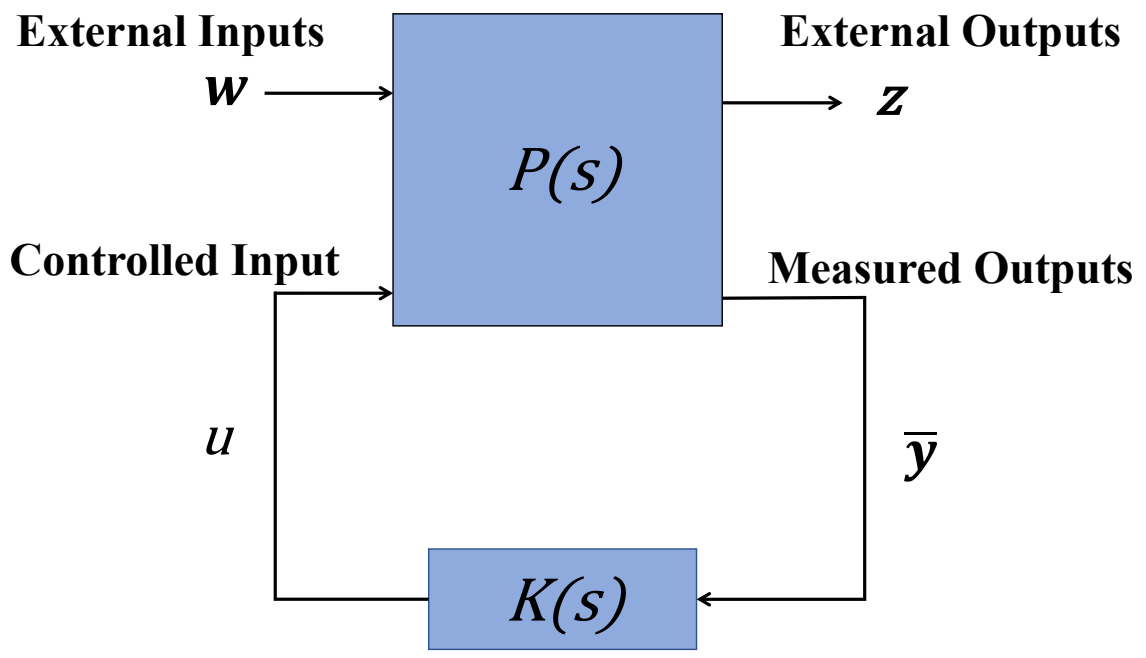

Fig. 1: LFT formulation of the $\mathcal{H}_\infty$ control problem.

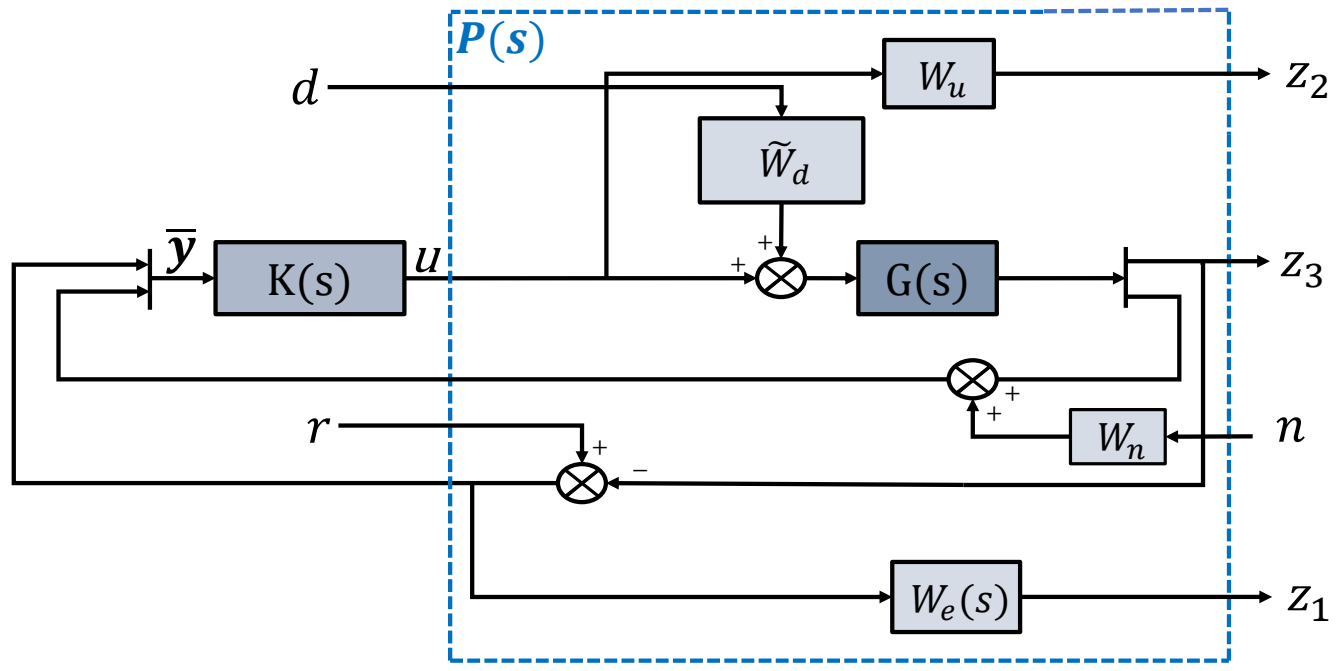

Fig. 2: Generalized plant for the considered problem.

Four main objectives are considered in this note: minimize reference tracking error, reduce the impact of disturbances and noises on tracking error, and avoid control input saturation. This results in the generalized plant $P(s)$ of Fig. 2, where $\boldsymbol{w} = [r, d, n]^T$, $\boldsymbol{z} = [z_1, z_2, z_3]^T$, $\bar{\boldsymbol{y}} = [(r-y), \dot{y}]^T$, and $y, \dot{y}$ are the system output and its derivative. $P(s)$ contains the open-loop dynamical system under consideration $G(s)$, as well as four weighting templates detailed below. Note that in the sequel, the sensitivity function between the input $w_i$ and the output $z_j$ corresponds to the transfer function between $w_i$ and $z_j$ excluding all weighting functions $W_\times$.

1) The transfer function $W_e(s) = \frac{\frac{s}{M_s}+\omega_b}{s+\omega_b\epsilon_e}$ shapes the sensitivity function $S(s)$ between the reference $r$ and the tracking error $z_1$. The primary objective is to minimize the closed-loop tracking error, mainly in steady state, by setting $\epsilon_e \ll 1$. In addition, $1/M_s$ specifies the desired modulus (or vector) margin [23]. A typical choice is $M_s = 2$, which ensures that the modulus margin is larger than 0.5 if $\gamma < 1$, i.e. that the gain and phase margins are larger than 6 dB and 30 deg respectively. Finally, $\omega_b$ defines the system's tracking speed.
2) The transfer function $W_d(s) = M_d W_e(s)$ shapes the sensitivity function $S_{d_i}(s)$ between the input disturbance $d$ and $z_1$ to reject external disturbances such as wind or gust, especially at low frequencies. $M_d \gg 1$ specifies the level of disturbance rejection. This amounts to choosing $\tilde{W}_d = M_d$ in Fig. 2. Note that this template may be modified by an additional weight to accommodate various disturbances and their respective frequency ranges.
3) The gain $W_u$ shapes the sensitivity function $KS(s)$ between $r$ and the control input $z_2$ to mitigate the risk of reaching actuator saturation.
4) The gain $W_n$ shapes the sensitivity function $S_n(s)$ between the output disturbance $n$ and the system's output $z_3$ to attenuate the impact of measurement noises, notably at high frequencies.

Classically, an optimal $\mathcal{H}_\infty$ controller is computed, which minimizes the value of $\gamma$ such that $||T_{\boldsymbol{w}\to\boldsymbol{z}}(s)||_\infty \leq \gamma$. It can be obtained by solving either Linear Matrix Inequalities (LMI) or algebraic Riccati equations, as discussed in [24]. Nevertheless, it is important to note that full-order controllers are obtained, in the sense that their order is equal to that of $P(s)$, which is usually high. Moreover, the following relation always holds:

$$\|T_{\boldsymbol{w}\to\boldsymbol{z}}(s)\|_\infty \leq \gamma \Rightarrow \begin{cases} \|T_{r\to z_1}(s)\|_\infty = \|W_e(s)S(s)\|_\infty \leq \gamma \\ \|T_{d\to z_1}(s)\|_\infty = \|W_d(s)S_{d_i}(s)\|_\infty \leq \gamma \\ \|T_{r\to z_2}(s)\|_\infty = \|W_u KS(s)\|_\infty \leq \gamma \\ \|T_{n\to z_3}(s)\|_\infty = \|W_n S_n(s)\|_\infty \leq \gamma \end{cases} \quad (5)$$

but the converse if usually not true. So minimizing a single transfer function between all exogenous inputs and outputs does not necessary lead to the lowest possible value of $\gamma$, which can potentially impact the overall system performance. These two problems are tackled by solving the $\mathcal{H}_\infty$ control problem using a non-smooth optimization technique, which makes it possible to freely choose the structure and the order of the controller [25], and to directly minimize $\gamma$ in the right-hand side of equation (5). In practice, this is achieved using MATLAB's **systune** function.

## III. Dynamical Modeling

We consider a 6-DoF (Degrees of Freedom) rigid quadcopter shown in Fig. 3, assumed to be symmetric, with a mass $m$ and an inertia matrix $I_B = \text{diag}(I_{xx}, I_{yy}, I_{zz})$. Two reference frames are defined: the world (or inertial) frame $\mathcal{F}_W$ with NED (North East Down) convention having basis vectors $(\boldsymbol{i_x}, \boldsymbol{i_y}, \boldsymbol{i_z})$, and the body frame $\mathcal{F}_B$ characterized by its axes $(x_B, y_B, z_B)$. The rotation from the body frame to the world frame is described by the Euler angles $\boldsymbol{\mu} = [\phi, \theta, \psi]^T$ through a rotation matrix $\mathcal{R}_B^W(\boldsymbol{\mu}) = [\boldsymbol{b_x}\ \boldsymbol{b_y}\ \boldsymbol{b_z}] \in \mathcal{SO}(3)$. The order of the axes used to get this rotation matrix follows the $ZYX$ convention. The drone's position in the world frame is represented by $\boldsymbol{\xi} = [x, y, z]^T$, and its velocity and acceleration are denoted by $\boldsymbol{v} = [v_x, v_y, v_z]^T$ and $\boldsymbol{a} = [a_x, a_y, a_z]^T$, respectively. In the body frame, we define the angular rates as $\boldsymbol{\Omega} = [p, q, r]^T$ and the angular accelerations as $\dot{\boldsymbol{\Omega}} = [\dot{p}, \dot{q}, \dot{r}]^T$.

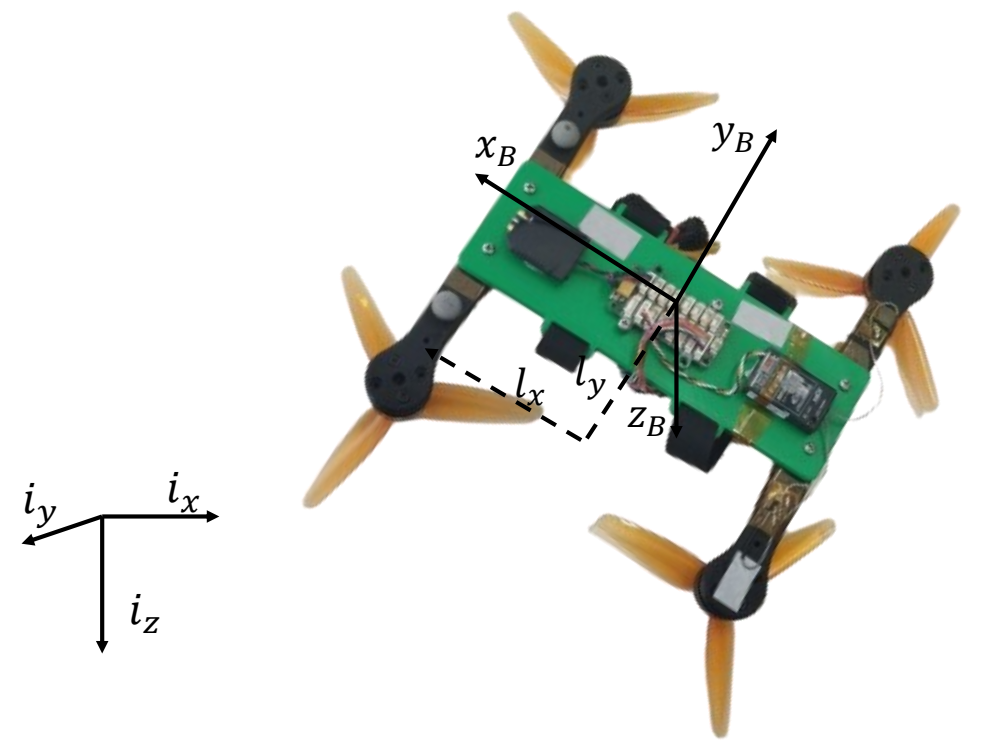

Fig. 3: Customized quadcopter built by the ENAC team and used in the experiments.

Applying Newton's Euler equations, the translational dynamics of the drone are given in the world frame by:

$$\dot{\boldsymbol{\xi}} = \boldsymbol{v} \qquad , \qquad \dot{\boldsymbol{v}} = \boldsymbol{a} = g\boldsymbol{i_z} + \frac{1}{m}(f_B \boldsymbol{b_z} + \boldsymbol{f_d}) \tag{6}$$

where $g\boldsymbol{i_z}$ represents the gravitational acceleration vector, and $\boldsymbol{f_d}$ includes external disturbance forces impacting the drone, such as drag or forces stemming from sudden gusts of wind. It is assumed that the propeller rotation axes are all parallel to the $z_B$ axis. The total scalar thrust $f_B$ they generate is therefore always aligned with $z_B$, and $f_B\boldsymbol{b_z}$ denotes the projection of $f_B$ in the world frame. The rotational dynamics, formulated in the body frame, are then given by:

$$\dot{\boldsymbol{\Omega}} = I_B^{-1}(\boldsymbol{\tau_B} - \boldsymbol{\Omega} \times I_B \boldsymbol{\Omega} + \boldsymbol{\tau_d}) \tag{7}$$

where $\boldsymbol{\tau_B}$ represents the total torque exerted by the propellers around the drone's axes, and $\boldsymbol{\tau_d}$ is a torque representing external disturbances. The vector torque $\boldsymbol{\tau_B}$ and the scalar thrust $f_B$ are linked to the angular velocities of the motors $\boldsymbol{\omega}$ through the following relation [12]:

$$\begin{bmatrix} \boldsymbol{\tau_B} \\ f_B \end{bmatrix} = \begin{bmatrix} I_B & 0_{3\times1} \\ 0_{1\times3} & m \end{bmatrix} \left[\tfrac{1}{2} G_1 \boldsymbol{\omega}^{\circ 2} + T_s G_2 \dot{\boldsymbol{\omega}}\right] \tag{8}$$

$$\text{where} \quad \left(\boldsymbol{\omega}^{\circ 2}\right)^T = \begin{bmatrix} \omega_1^2 & \omega_2^2 & \omega_3^2 & \omega_4^2 \end{bmatrix}^T$$

$T_s$ is the sampling time introduced here to simplify future calculations. $G_1, G_2 \in \mathbb{R}^{4\times4}$ are matrices that quantify the impact of each rotor on the drone's acceleration, showing the propellers' effectiveness. They are expressed as follows:

$$\begin{aligned} G_1 &= \begin{bmatrix} I_B^{-1} & 0_{3\times1} \\ 0_{1\times3} & \frac{1}{m} \end{bmatrix} \begin{bmatrix} -l_y K_\tau & l_y K_\tau & l_y K_\tau & -l_y K_\tau \\ l_x K_\tau & l_x K_\tau & -l_x K_\tau & -l_x K_\tau \\ K_q & -K_q & K_q & -K_q \\ -K_\tau & -K_\tau & -K_\tau & -K_\tau \end{bmatrix} \\ G_2 &= T_s^{-1} \begin{bmatrix} 0 & 0 & 0 & 0 \\ 0 & 0 & 0 & 0 \\ I_{r_{zz}} & -I_{r_{zz}} & I_{r_{zz}} & -I_{r_{zz}} \\ 0 & 0 & 0 & 0 \end{bmatrix} \end{aligned} \tag{9}$$

where $l_x$ and $l_y$ represent the distances from the drone's center to its motors on the $x_B$ and $y_B$ axes respectively, as illustrated in Fig. 3, $K_\tau$ and $K_q$ denote the thrust coefficient of each propeller and the drag coefficient due to its rotation respectively, and $I_{r_{zz}}$ is the moment of inertia of both a motor and a propeller. The matrix $G_2$ is crucial for modeling the gyroscopic torque effects resulting from the rotation of the motors and propellers. Note that the inertia matrix and the mass are added in (9) to express the control effectiveness matrices in terms of accelerations.

## IV. INDI/$\mathcal{H}_\infty$ Cascaded Control Architecture

The proposed control architecture is shown in Fig. 4, where the subscripts $ref$, $c$, $m$ denote reference, commanded and measured values respectively. It consists of an inner stabilization loop and an outer guidance loop, each involving a combination of INDI and linear $\mathcal{H}_\infty$ controllers. The latter first generate the commanded accelerations $\boldsymbol{\nu_{\dot{\Omega}_c}}$ and $\boldsymbol{\nu_{a_c}}$ for the attitude and position dynamics respectively, which are then sent to the INDI controllers. The design of these two loops is thoroughly described in this section, including a comparison with an existing INDI/PD approach to show the ability of the proposed architecture to improve the system's robustness to gust-type perturbations. It is worth mentioning that in the presence of a trajectory generator, which is outside the scope of this note, the proposed cascaded architecture can be used to track both position $(\boldsymbol{\xi_{ref}}, \boldsymbol{\mu_{ref}})$ and velocity $(\boldsymbol{v_{ref}}, \boldsymbol{\Omega_{ref}})$. In what follows we focus on tracking accurately the drone's position $\boldsymbol{\xi_{ref}}$ in the presence of disturbances.

### A. INDI/$\mathcal{H}_\infty$ Stabilization Loop

*1) Inner-INDI Control Law:* Based on the theoretical explanations presented in Sections II-A and II-B, the proposed INDI/$\mathcal{H}_\infty$ methodology to control the drone's rotational dynamics is now presented. In general, the gyroscope sensor can measure the angular rates $\boldsymbol{\Omega}$, however the rotational dynamics in equation (7) involve the angular accelerations $\dot{\boldsymbol{\Omega}}$. To be able to apply the INDI linearization technique, it is therefore essential to have knowledge of the angular accelerations. To address this issue, the latter are directly computed from the measured angular rates. It is indeed shown by [26] that applying a second order filter $H(s)$ before performing this derivation is sufficient to reduce noises of the gyroscope sensor. Due to the incremental nature of the controller, synchronizing the incremented signals with the measured data is essential, and for that reason the same filter should be used in both INDI guidance and stabilization loops [13]. In the sequel, all the filtered measured data are presented in the equations with the subscript $f$. Starting from equation (7), the INDI controller is designed by assuming that the external disturbances $\boldsymbol{\tau_d}$ can be estimated using filtered measurements:

$$\boldsymbol{\tau_d} = I_B \dot{\boldsymbol{\Omega}}_f - \boldsymbol{\tau_{B_f}} + \boldsymbol{\Omega_f} \times I_B \boldsymbol{\Omega_f} \tag{10}$$

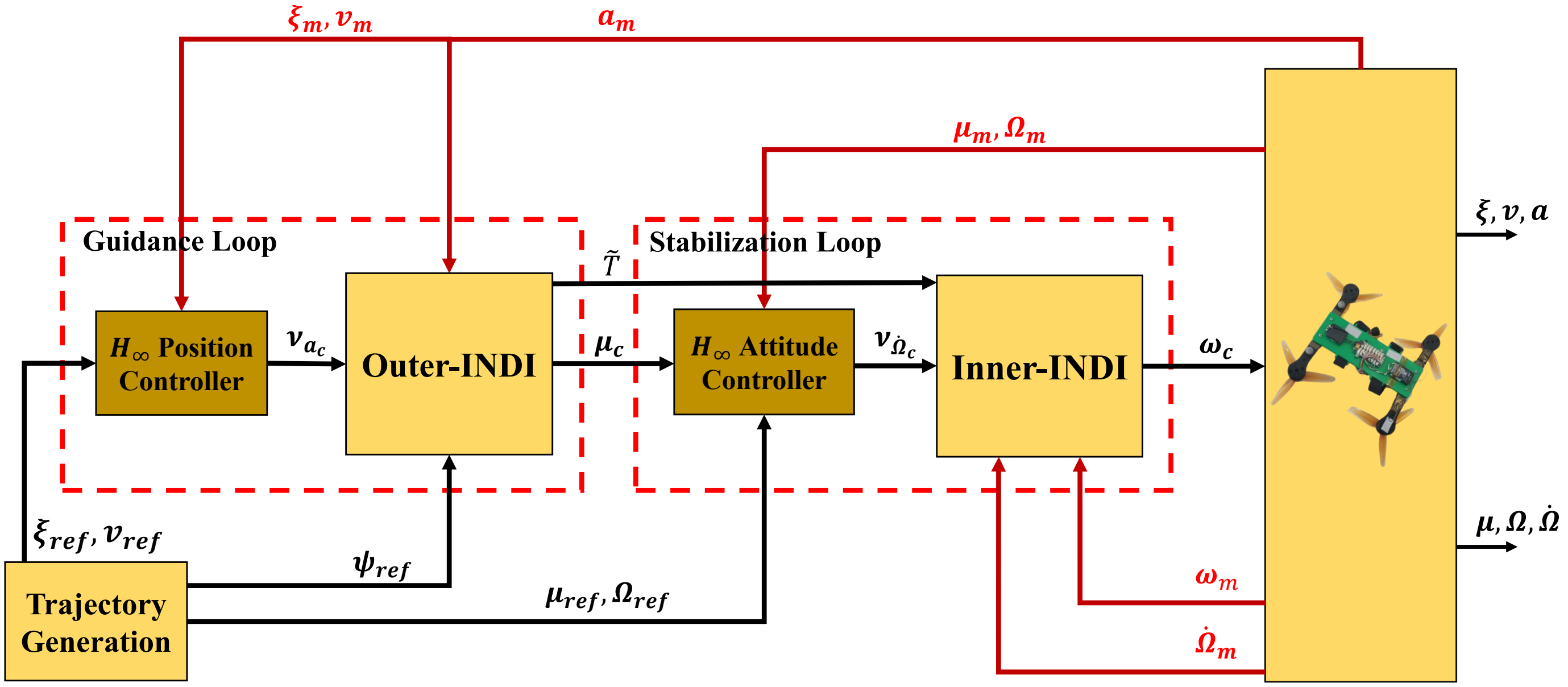

Fig. 4: Proposed cascaded control architecture using a mixed INDI/$\mathcal{H}_\infty$ controller for each loop.

where $\boldsymbol{\tau_{B_f}}$ is obtained from equation (III) by measuring or estimating $\boldsymbol{\omega}$, see Section V-A. By combining equations (10) and (7), the following relation is obtained:

$$I_B(\dot{\boldsymbol{\Omega}} - \dot{\boldsymbol{\Omega}}_{\boldsymbol{f}}) = (\boldsymbol{\tau_B} - \boldsymbol{\tau_{B_f}}) + (\boldsymbol{\Omega_f} \times I_B \boldsymbol{\Omega_f} - \boldsymbol{\Omega} \times I_B \boldsymbol{\Omega}) \quad (11)$$

The difference between the gyroscopic angular momentum and the filtered one – last term in the equation (11) – is usually small and can be considered acting as external disturbances. After linearizing equation (III) by applying a Taylor expansion as done in [12], equation (11) can then be rewritten in terms of $G_1$ and $G_2$:

$$\begin{bmatrix} \dot{\boldsymbol{\Omega}} - \dot{\boldsymbol{\Omega}}_{\boldsymbol{f}} \\ \frac{1}{m}\tilde{T} \end{bmatrix} = G_1 \mathrm{diag}(\boldsymbol{\omega_f})(\boldsymbol{\omega} - \boldsymbol{\omega_f}) + T_s G_2 (\dot{\boldsymbol{\omega}} - \dot{\boldsymbol{\omega}}_{\boldsymbol{f}}) \quad (12)$$

$\tilde{T}$ is the thrust control increment calculated by the guidance controller, as shown in Fig. 4, and it is divided by mass to maintain acceleration units. The variation in motor angular speed can be calculated in discrete time as $\dot{\boldsymbol{\omega}} = \frac{\boldsymbol{\omega} - L\boldsymbol{\omega}}{T_s}$, where $L$ stands for the lag operator. In addition, it is pointed out in [12] that modeling torques and forces linearly with respect to the rotational speed of the rotor, rather than quadratic, simplifies the computations and leads to a negligible error. Therefore, equation (12) becomes:

$$\begin{bmatrix} \dot{\boldsymbol{\Omega}} - \dot{\boldsymbol{\Omega}}_{\boldsymbol{f}} \\ \frac{1}{m}\tilde{T} \end{bmatrix} = (G_1 + G_2)(\boldsymbol{\omega} - \boldsymbol{\omega_f}) - G_2 L(\boldsymbol{\omega} - \boldsymbol{\omega_f}) \quad (13)$$

where $G_1$ and $G_2$ are no longer computed from equation (9), but should be estimated using measured data to minimize the approximation error between equations (12) and (13), see Section V-A. This equation is then inverted, finally leading to the following INDI control law, which computes the commanded motor angular velocity $\boldsymbol{\omega_c}$ sent to the actuators as a function of the commanded angular accelerations $\dot{\boldsymbol{\Omega}} = \nu_{\dot{\boldsymbol{\Omega}}_{\boldsymbol{c}}}$ coming from the $\mathcal{H}_\infty$-based attitude controller designed in Section IV-A2.

$$\boldsymbol{\omega_c} = \boldsymbol{\omega_f} + (G_1 + G_2)^\dagger \left( \begin{bmatrix} \boldsymbol{\nu}_{\dot{\boldsymbol{\Omega}}_{\boldsymbol{c}}} - \dot{\boldsymbol{\Omega}}_{\boldsymbol{f}} \\ \frac{1}{m}\tilde{T} \end{bmatrix} + G_2 L(\boldsymbol{\omega_c} - \boldsymbol{\omega_f}) \right) \quad (14)$$

*2) $\mathcal{H}_\infty$-Based Attitude Controller:* When applying the INDI control law (14), the transfer function between $\boldsymbol{\nu}_{\dot{\boldsymbol{\Omega}}_{\boldsymbol{c}}}$ and $\dot{\boldsymbol{\Omega}}$ is reduced to the actuator dynamics $A(s)$ [12], usually modeled as a first order low pass filter. The dynamics of the loop which encompasses the inner-INDI control law designed above and the dynamics of the drone – denoted by $G(s)$ in the sequel as in Section II-B – are therefore pretty simple: they are only composed of the actuator dynamics and double integrators, as shown in Fig. 5 for the $\phi$ channel (note that the remainder of Section IV-A focuses on the $\phi$ channel, but similar results are obtained for the $\theta$ and $\psi$ channels). An $\mathcal{H}_\infty$ attitude controller $K(s)$ is now designed to generate the virtual control input $\boldsymbol{\nu}_{\dot{\boldsymbol{\Omega}}_{\boldsymbol{c}}}$ sent to $G(s)$, using the control framework of Fig. 5 and the weighting templates of Table I. Note that $W_n$ is quite low, which means that the noise channel is almost not penalized. Indeed, the filtering of $\boldsymbol{\Omega}$ by $H(s)$ already reduces noise sufficiently. The number of states of $G(s)$ is 3, and with a first order dynamic template $W_e(s)$, the generalized plant $P(s)$ is of order 4. The resulting full-order controller is therefore of order 4. However, it appears that two of its poles are fast and can be eliminated. A reduced-order controller of order 2 is then designed instead by following the methodology of Section II-B. It is structured into two distinct first order controllers $K_\mu(s)$ and $K_\Omega(s)$, which use the available measures of $\phi$ and $p$ respectively. The resulting two-input single-output controller $K(s)$ is structured in a cascaded way, i.e. $K(s) = K_\Omega(s)[K_\mu(s) \quad -1]$. It is used here to track a position command $\boldsymbol{\mu_c}$ coming from the guidance loop, but this particular structure would also allow to track both position and velocity references $(\boldsymbol{\mu_{ref}}, \boldsymbol{\Omega_{ref}})$ coming from a trajectory generator, as shown in Fig. 4.

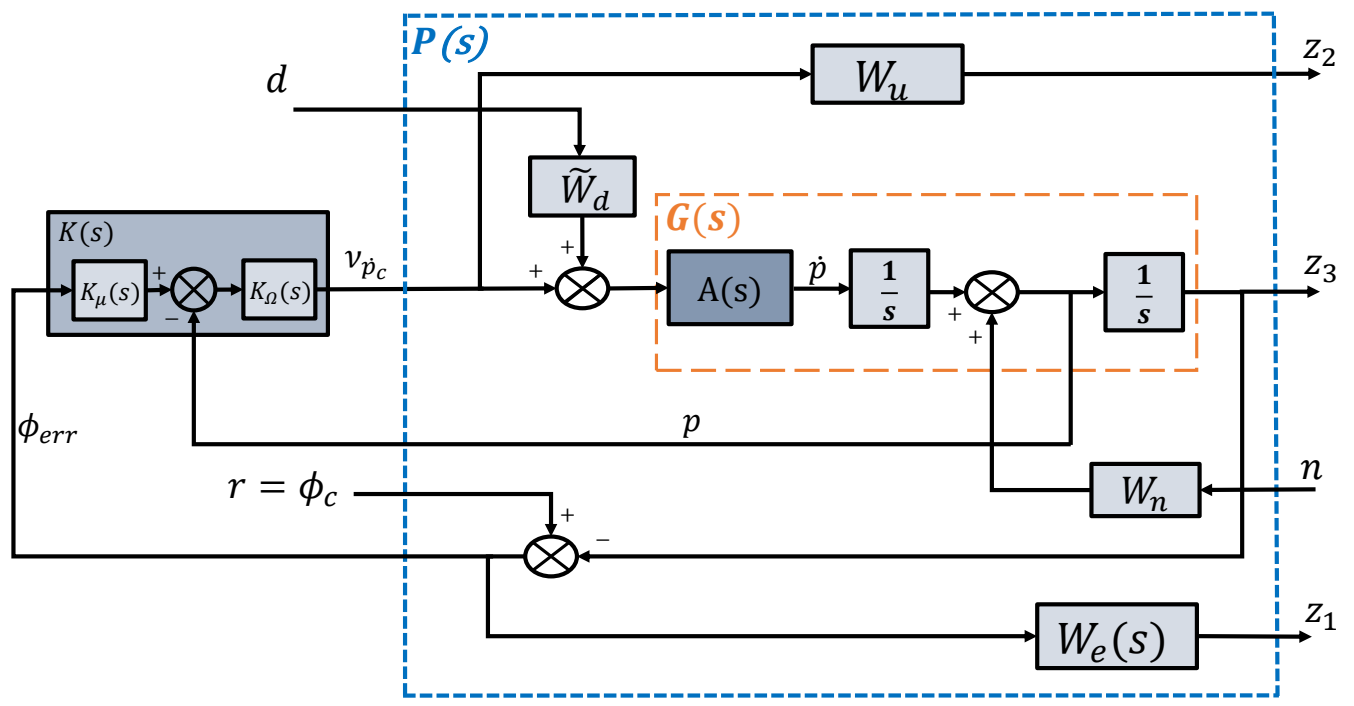

Fig. 5: Inner loop $\mathcal{H}_\infty$ control problem formulation for the rotational dynamics.

TABLE I: Weighting templates for the $\mathcal{H}_\infty$ attitude controller

| | |
|---|---|
| $W_e(s)$ | $\dfrac{\frac{s}{2} + 2\pi 1.8}{s + 2\pi 1.8 \times 0.0001}$ |
| $W_u$ | 800 |
| $W_d(s)$ | $400\, W_e(s)$ |
| $W_n$ | 0.1 |

*3) Sensitivity Analysis:* To validate the $\mathcal{H}_\infty$ controllers and demonstrate the improvement of the INDI/$\mathcal{H}_\infty$ approach over the standard INDI/PD one, a comparison is conducted for a Bebop quadcopter model using the same parameters as those given in [13]. The actuator model $A(s)$ of equation (15) is used, with a time constant $\tau_m = \frac{1}{53.94}$ s, that leads to the same response as the discrete time actuator of [13], and the chosen second order low pass filter $H(s)$ has damping $\xi_n = 0.55$ and natural frequency $\omega_n = 50$ rad/s:

$$A(s) = \frac{1}{\tau_m s + 1} \qquad \text{and} \qquad H(s) = \frac{\omega_n^2}{s^2 + 2\xi_n \omega_n s + \omega_n^2} \tag{15}$$

A PD controller with the same cascaded structure as the $\mathcal{H}_\infty$ controllers of Section IV-A2 is considered, whose gains $K_\mu^{PD} = 10.7$, $K_\Omega^{PD} = 28$ are taken from [13]. In order to have a fair comparison between all controllers, it is worth noting that the weighting templates $W_e(s)$ and $W_u$ in Table I have been chosen so that the $\mathcal{H}_\infty$ controllers have a similar response as the PD controller in terms of rise time, settling time and steady-state error (see Fig. 7a). This allows to specifically highlight the improvement in disturbance attenuation (see below), but it should be emphasized that these templates could be further optimized for a faster response, taking care of avoiding actuator saturation.

A comparative sensitivity analysis of the three attitude controllers (INDI/PD, full-order INDI/$\mathcal{H}_\infty$ and reduced-order INDI/$\mathcal{H}_\infty$) is presented in Fig. 6, based on the closed-loop architecture of Fig. 5. The system's sensitivity $S(s)$ is almost the same and below the defined template in all cases, with an approximate bandwidth frequency of $\omega_T \approx 9$rad/s. The control sensitivity $KS(s)$ shows that the designed INDI/$\mathcal{H}_\infty$ controllers stay within the actuator's physical saturation limits. Then we make a difference between input and output disturbances, which affect the commanded control input $\nu_{\dot{p}_c}$ and the angular acceleration $\dot{p}$ respectively. The former correspond to physical uncertainties that can disturb the behavior of the drone, such as gust or wind. They are effectively minimized during the $\mathcal{H}_\infty$ design process, whereas the latter are generally taken into account when computing the INDI control law. In both cases, there is a notable difference between the three controllers at low frequencies, where it is known that disturbances usually appear. The full-order INDI/$\mathcal{H}_\infty$ controller demonstrates superior disturbance attenuation, with a marginal advantage over the reduced-order one, while the INDI/PD controller shows a significant deviation from the performance of the other two controllers. In addition, comparing the sensitivities $S_{d_i}(s)$ and $S_{d_o}(s)$ (corresponding to the input and output disturbances) of the three controllers shows that output disturbances are better rejected, due to the estimation of the disturbance torque $\tau_d$ by the INDI control law. Overall, the results highlight the effectiveness of the proposed INDI/$\mathcal{H}_\infty$ approach in enhancing disturbance rejection capabilities, particularly at low frequency, without exceeding the actuation limits. It also shows the importance of the disturbance estimation in the INDI control law in enhancing the system's robustness.

*4) Simulation-Based Validation:* Figure 7a illustrates the tracking performance of the three controllers, which is quite similar when following a step input reference for the roll angle $\phi$. The rising time of the system is about 0.2 s in all cases, with a very small overshoot for the reduced INDI/$\mathcal{H}_\infty$ controller. The capability of the controllers to attenuate disturbances is then evaluated in a hovering scenario. The angles are initially set to zero, and the drone encounters some disturbances. Figure 7b clearly indicates that, compared with the standard INDI/PD controller, the disturbance attenuation is improved by nearly 50% with the reduced-order INDI/$\mathcal{H}_\infty$ controller and by more than 70% with the full-order INDI/$\mathcal{H}_\infty$ controller. These results support the aforementioned sensitivity analysis.

### *B. INDI/$\mathcal{H}_\infty$ Guidance Loop*

*1) Outer-INDI Control Law:* The translational acceleration measurements $\boldsymbol{a_m}$ are mainly derived from a fusion of different sensors, like Inertial Measurement Units (IMU) and Global Positioning System (GPS). It is usually presented in the $\mathcal{F}_w$ frame to enhance the system's capability

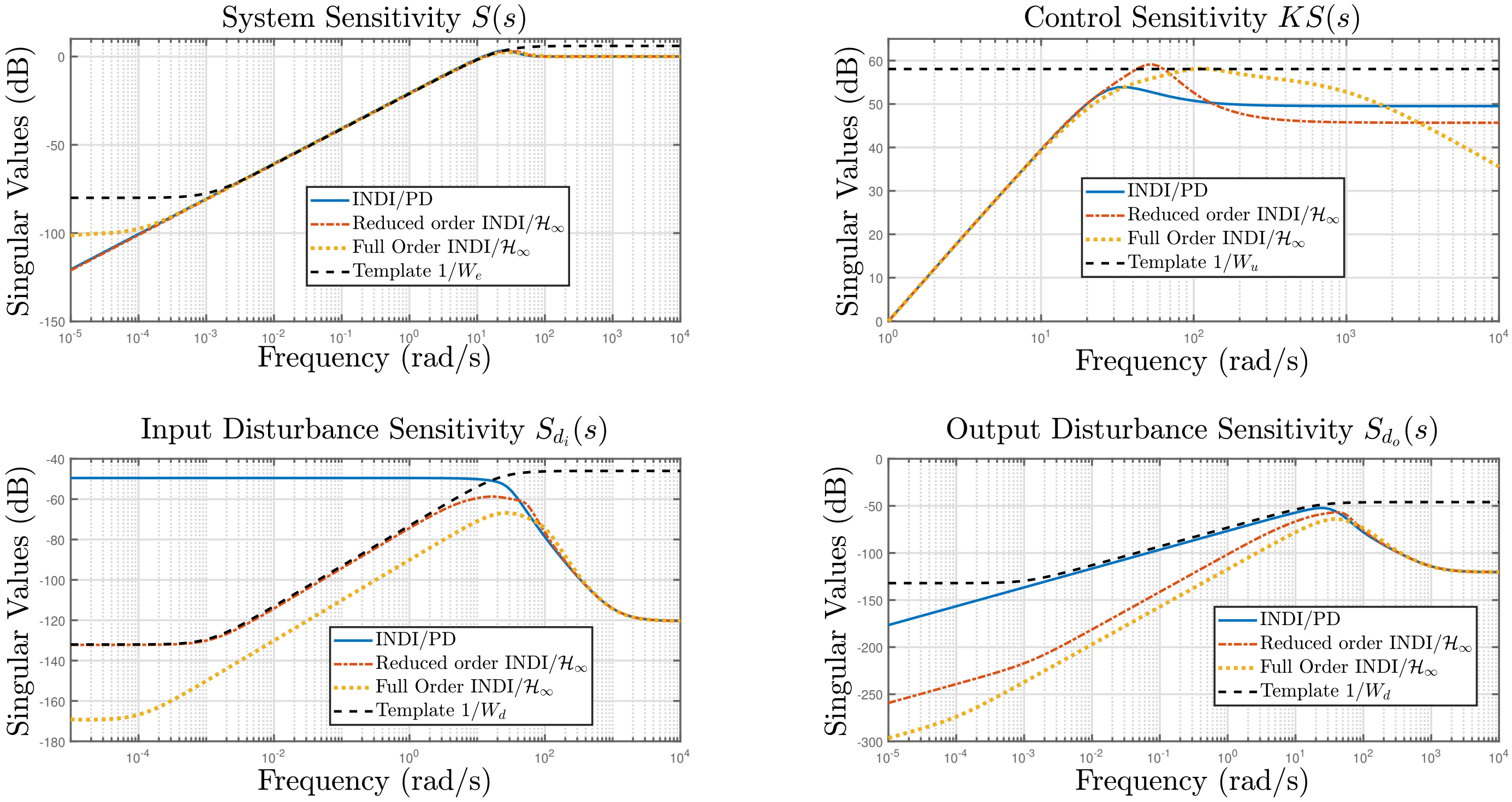

Fig. 6: Frequency analysis comparison of the three attitude controllers.

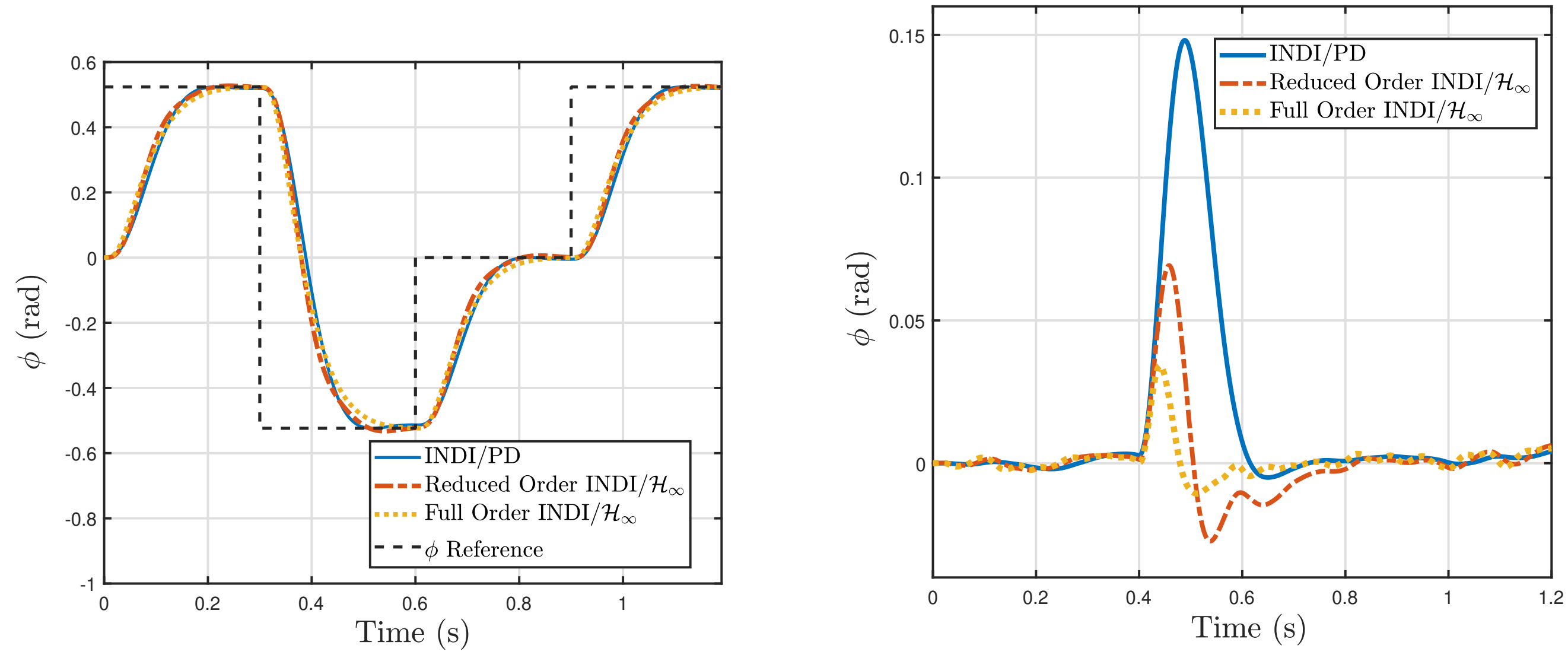

(a) Roll angle tracking

(b) Roll angle in the presence of a disturbance

Fig. 7: Attitude controller performance comparison during tracking (a) and disturbance rejection (b).

to accurately follow a trajectory generated in the $\mathcal{F}_w$ frame. Therefore, the INDI control law is formulated in the same frame. It tries to align the filtered acceleration $\boldsymbol{a_f}$, obtained from $\boldsymbol{a_m}$ using the same filter $H(s)$ as in Section IV-A1, with the commanded ones $\boldsymbol{\nu_{a_c}}$, generated by the $\mathcal{H}_\infty$ position controller. By reformulating equation (6), the external disturbance forces $\boldsymbol{f_d}$ can be expressed as follows:

$$\boldsymbol{f_d} = m\boldsymbol{a_f} - mg\boldsymbol{i_z} - f_{B_f}\boldsymbol{b_{z_f}} \tag{16}$$

where $f_{B_f}$ is the filtered scalar thrust obtained from equation (III) by measuring or estimating $\boldsymbol{\omega}$, and $\boldsymbol{b_{z_\times}}$ is the unit vector of the vertical axis of $\mathcal{F}_B$ expressed in $\mathcal{F}_W$ given as :

$$\boldsymbol{b_{z_\times}} = \begin{bmatrix} s_{\phi_\times} s_{\psi_\times} + c_{\phi_\times} c_{\psi_\times} s_{\theta_\times} \\ c_{\phi_\times} s_{\psi_\times} s_{\theta_\times} - c_{\psi_\times} s_{\phi_\times} \\ c_{\phi_\times} c_{\theta_\times} \end{bmatrix}^T \tag{17}$$

where $s_\star = \sin(\star)$, $c_\star = \cos(\star)$, and the subscript $\times$ is either $f$ or $c$. By substituting equation (16) into equation (6), the incremental form of the outer-INDI controller can be computed as:

$$f_{B_c}\boldsymbol{b_{z_c}} = m(\boldsymbol{\nu_{a_c}} - \boldsymbol{a_f}) + f_{B_f}\boldsymbol{b_{z_f}} \tag{18}$$

where $f_{B_c}$ is the desired scalar thrust and $\boldsymbol{b_{z_c}}$ is expressed in term of the desired Euler angles $\boldsymbol{\mu_c} = [\phi_c, \theta_c, \psi_c]^T$. The thrust control increments used in equations (12,13,14) is then given by:

$$\tilde{T} = \left\| f_{B_c}\boldsymbol{b_{z_c}} - f_{B_f}\boldsymbol{b_{z_f}} \right\| \tag{19}$$

The commanded heading $\psi_c = \psi_{ref}$ directly comes from the desired trajectory. The other control inputs sent to the stabilization loop are $f_{B_c}, \phi_c$ and $\theta_c$. They are computed from $f_{B_c}\boldsymbol{b_{z_c}}$, which is itself obtained from equation (18), using equation (20).

$$f_{B_c}\boldsymbol{b_{z_c}} = \begin{bmatrix} T_{x_c} \\ T_{y_c} \\ T_{z_c} \end{bmatrix} \Rightarrow \begin{cases} f_{B_c} = \|f_{B_c}\boldsymbol{b_{z_c}}\| \\ \phi_c = \arcsin\left(\frac{\sin(\psi_c)T_{x_c} - \cos(\psi_c)T_{y_c}}{f_{B_c}}\right) \\ \theta_c = \arcsin\left(\frac{\cos(\psi_c)T_{x_c} + \sin(\psi_c)T_{y_c}}{f_{B_c}\cos(\phi_c)}\right) \end{cases} \tag{20}$$

*2) $\mathcal{H}_\infty$-Based Position Controller:* A robust controller is then designed to generate the commanded acceleration $\boldsymbol{\nu_{a_c}}$, which acts as the virtual control input to the outer-INDI controller. The system in equation (6) is linearized by applying Taylor's expansion around the hovering position. The linearized model is given in equation (21). Considering the coupling between the dynamics of the drone, it is important to consider the closed stabilization loop when designing the linear controller for the guidance loop. Incorporating the stabilization loop ensures the achievement of well-damped poles for the overall system and prevents the outer loop from being designed on faster frequencies than the inner loop. This, for sure, can increase the order of the control law when using a full-order $\mathcal{H}_\infty$ controller. However, it can then be reduced by using structured $\mathcal{H}_\infty$ design tools.

$$\dot{\boldsymbol{v}} = \begin{bmatrix} a_x \\ a_y \\ a_z \end{bmatrix} = \begin{bmatrix} 0 & -g & 0 \\ g & 0 & 0 \\ 0 & 0 & \frac{1}{m} \end{bmatrix} \begin{bmatrix} \phi \\ \theta \\ f_B \end{bmatrix} \tag{21}$$

The $\mathcal{H}_\infty$ control problem is formulated as shown in Fig. 8, where $G(s)$ in this case contains double integrators, $H(s)$ filters, and the stabilization loop. It is presented here for the translational dynamics in the $x$ axis only, since it is similar for the other axes $(y, z)$. The controller $K(s)$ generates the acceleration $\nu_{a_{x_c}}$, which is then used as a virtual commanded input for the INDI control law described previously in equation (18). The plant $G(s)$ in Fig. 8 is of order 11, having two integrators, two second order filters $H(s)$, and the stabilization loop of order 5 obtained in Section IV-A. The values of the tuned templates are presented in Table II. Classically solving the $\mathcal{H}_\infty$ control problem using the LMI approach results in a controller of order 12, the same order as the generalized plant $P(s)$. Then the control law is structured and reduced to a $2^{nd}$ order controller using the same process as described in Section II-B.

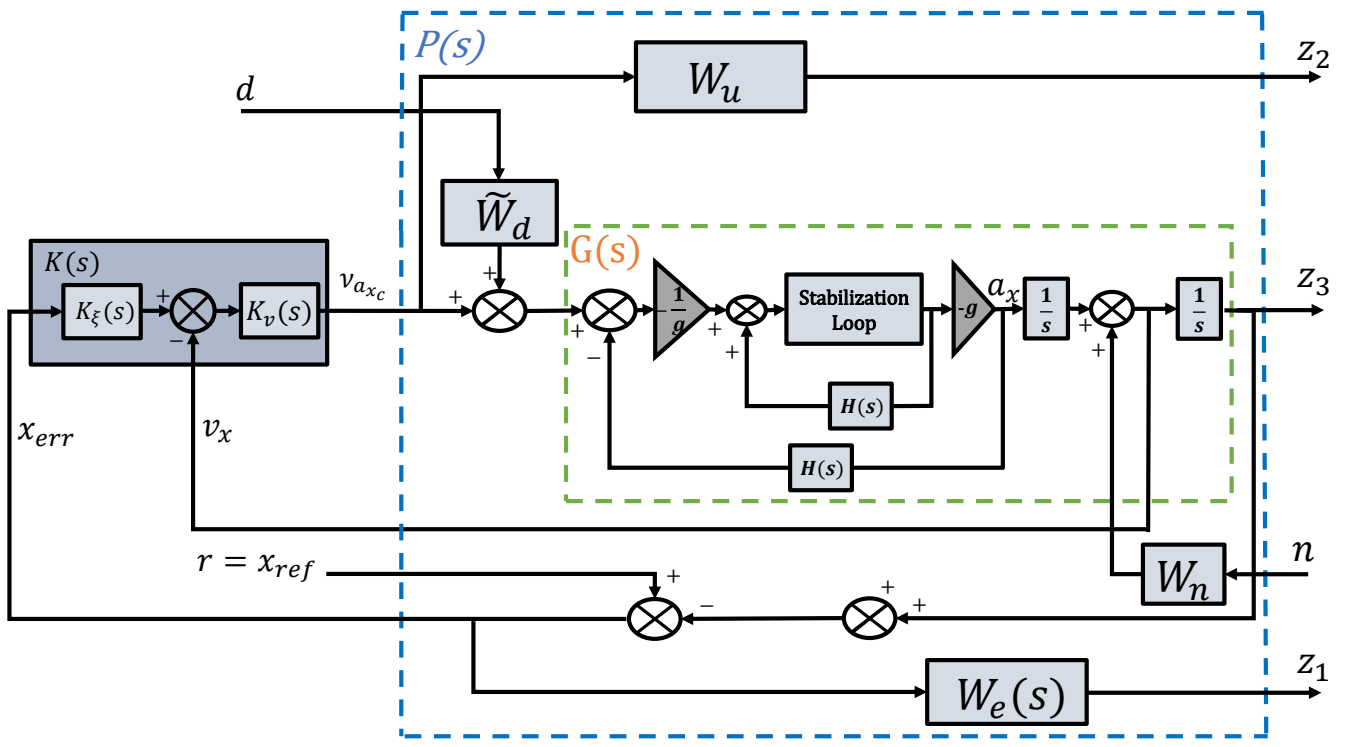

Fig. 8: Outer loop $\mathcal{H}_\infty$ control problem formulation for the translational dynamics.

TABLE II: Weighting templates for the $\mathcal{H}_\infty$ position controller

| | |
|---|---|
| $W_e(s)$ | $\frac{\frac{s}{2} + 2\pi 0.1}{s + 2\pi 0.1 \times 0.0001}$ |
| $W_u$ | 1.2 |
| $W_d(s)$ | $10\, W_e(s)$ |
| $W_n$ | 4 |

*3) Sensitivity Analysis:* The weighting templates given in Table II have been tuned to have almost the same closed-loop system's nominal response as with the PD controller tuned by [13]. The latter has the same cascaded control architecture, with gains $K_\xi^{PD} = 0.7$ and $K_v^{PD} =$ 1.5. The sensitivity analysis of the designed controllers is shown in Fig. 9. It is observed that the system's sensitivity $S(s)$ is very similar in all cases. In particular, there is no steady-state error, and the bandwidth frequency is the same. The control sensitivity $KS(s)$ is below the defined template in the entire frequency range for the three controllers, which shows that the physical limitations are not exceeded. Then, as in Section IV-A3, we make a difference between input and output disturbances that affects $\nu_{a_{x_c}}$ and $a_x$ respectively. It is observed that the input disturbance sensitivity $S_{d_i}(s)$ and the output disturbance sensitivity $S_{d_o}(s)$ are both better with the INDI/$\mathcal{H}_\infty$ controllers than with the INDI/PD controller at low frequencies, as shown in the bottom plots of Fig. 9, indicating the superiority of the INDI/$\mathcal{H}_\infty$ approach in attenuating disturbances. By comparing the disturbance sensitivities $S_{d_i}(s)$ and $S_{d_o}(s)$ of the INDI/PD, it is evident that the INDI controller is effective in improving robustness to output disturbances. A similar conclusion can be drawn by examining the disturbance sensitivities of the INDI/$\mathcal{H}_\infty$. But the INDI/$\mathcal{H}_\infty$ controllers show an additional significant improvement in disturbance attenuation over the classical INDI/PD controller for both input and output disturbances. Therefore, having a robust linear controller clearly enhances the closed-loop's ability to attenuate disturbances.

*4) Simulation-Based Validation:* Figure 10a shows the drone's movement in the $x$ axis direction while flying through defined waypoints between $-2$m and 2m. As expected, the reduced-order INDI/$\mathcal{H}_\infty$ controller's performances are nearly identical to that of the INDI/PD controller in the nominal case, having almost the same rise time and overshoot. Additionally, the drone is subjected to some disturbances, modeled by step inputs of different amplitudes applied at different times. These disturbances are applied as an external forces acting on the drone with an amplitude varying between 1.5N and 6N. The system's ability to attenuate these disturbances is improved by about 70% when using the reduced-order INDI/$\mathcal{H}_\infty$ controller as shown in the box of Fig. 10a. It can also be seen in Fig. 10b that the pitch angle $\theta_c$ generated by both controllers is always below saturation ($45^\circ \approx 0.78$ rad), and is only slightly larger with the INDI/$\mathcal{H}_\infty$ controller than with the INDI/PD one.

## V. Experimental Validation

The proposed control architecture has been validated experimentally using a wind generator, as shown in Fig. 11. A customized quadcopter, built in-house at ENAC and shown in Fig. 3, is used for the experiments. It features a Paparazzi Tawaki v1.1 autopilot and runs Paparazzi software[1]. The tests are conducted in the *Volière Drones Toulouse-Occitanie* indoor flight arena[2], equipped with a motion capture system to localize the position of the drone. The primary purpose of the tests is to demonstrate the ability of the INDI/$\mathcal{H}_\infty$ controllers to enhance the system's robustness against external disturbances. The quadcopter's performance is therefore tested in two scenarios: first hovering in front of the wind generator to analyze the robustness with respect to various wind steps and then passing through predefined waypoints despite wind disturbances.

### A. System's Physical Estimation

It is possible to estimate the drone's physical parameters using classical system identification methods, but this process is time-consuming and not the main focus here. Instead, we use a simpler approach to obtain the necessary physical parameters. Initially, $G_1 + G_2$ and $G_2$ are manually tuned for basic flight capability. Then, flight tests are conducted to have better estimation of the control effectiveness matrices and actuator model. The matrices are expressed in terms of incremental accelerations, measured by onboard gyroscope and accelerometer, as shown in equation (13).

$$\begin{bmatrix} \Delta\dot{\boldsymbol{\Omega}}_{\boldsymbol{f}} \\ \Delta a_{z\,f} \end{bmatrix} = (G_1 + G_2)\Delta\boldsymbol{\omega_f} - G_2 L \Delta\boldsymbol{\omega_f} \tag{22}$$

where $\Delta$ represents the difference between two consecutive measurements with a frequency of 500Hz. It appears from the expression of $G_2$ in (9) that it only affects the yaw dynamics of the drone. Since we are not interested in estimating precisely the value of $G_2$, the yaw dynamics is set to zero in all the tests. Equation (22) is then simplified and the last term $G_2 L\Delta\boldsymbol{\omega_f}$ is eliminated. The second order filter $H(s)$ used in filtering all the data is defined in equation (15). Finally, one of the challenges with the used drone is the lack of measurements of the motor angular speed. However, it is possible to log the throttle commands (PWM, Pulse Width Module) sent by the autopilot to each motor. The actuator model is then used to estimate the motor angular speed using:

$$\hat{\omega} = \hat{A}(s)\, u_m \tag{23}$$

where $\hat{\omega}$ is the estimated motor angular speed, $\hat{A}(s)$ is the estimated model of the actuator given initially from the manufacturer data sheet, and $u_m$ is the throttle command sent to each motor. The estimated motor angular speed is then used in the simplified version of equation (22). After that, the estimated time constant $\hat{\tau}_m$ of $\hat{A}(s)$ presented in equation (15), and the estimated control effectiveness matrix $\hat{G}_{12}$ that replace $(G_1 + G_2)$ in equation (22) are obtained by solving the following optimization problem using MATLAB's **fmincon** function:

$$\min_{\Phi} \|y - \hat{y}(\Phi)\|_2 \text{ where } \Phi = \{\hat{\tau}_m, \hat{G}_{12}\} \tag{24}$$

where $\boldsymbol{y} = [\Delta\dot{\boldsymbol{\Omega}}_{\boldsymbol{f}}{}^T, \Delta a_{z_f}{}^T]^T$ and $\hat{y}(\Phi) = [\Delta\hat{\dot{\boldsymbol{\Omega}}}_{\boldsymbol{f}}{}^T, \Delta\hat{a}_{z_f}^T]^T$ is the estimated accelerations computed using equations (22-23). The initialization values of the optimization problem are the ones obtained by solving analytically equation (22) using Moore-Penrose pseudo-inverse. The estimated actuator transfer function $\hat{A}(s)$ is validated using logged commanded and measured angular accelerations. The fitting percentage between the measured data and the simulated data is 37.5% and the estimated time constant is $\hat{\tau}_m = \frac{1}{25.65}$ s. It is clear then, within all the assumptions considered to obtain $\hat{\tau}_m$ and $\hat{G}_{12}$, that the estimated values include different uncertainties. Due to that, the inversion model in each INDI controller is obviously not accurate. However, the estimated values will be used to tune both the $\mathcal{H}_\infty$ and the PD controllers, demonstrating that the $\mathcal{H}_\infty$ controller can effectively handle modeling uncertainties while maintaining good performances.

[1]https://wiki.paparazziuav.org

[2]https://www.enac.fr/en/drone-flight-arena-toulouse-occitanie-0

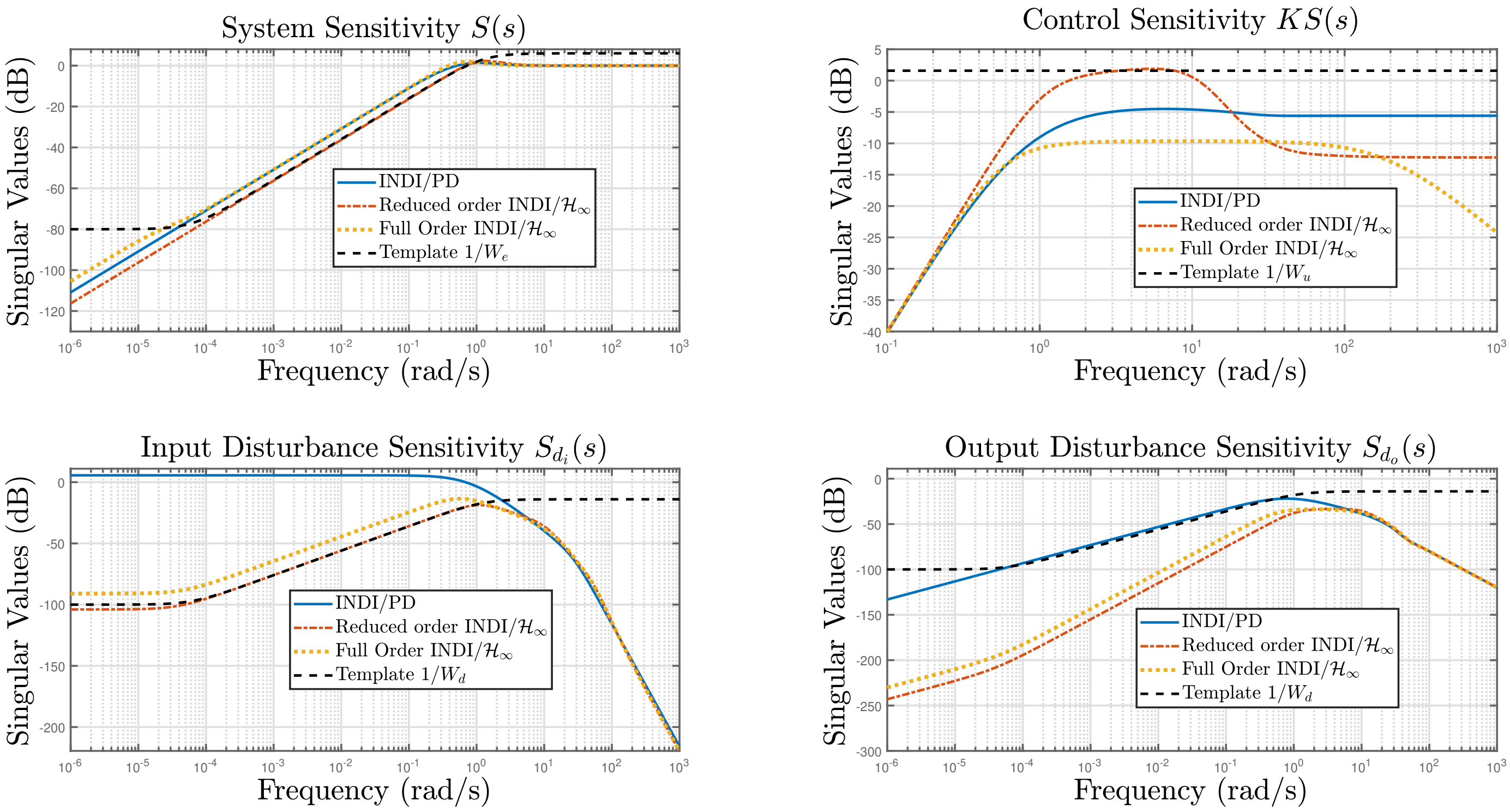

Fig. 9: Frequency analysis comparison of the three position controllers.

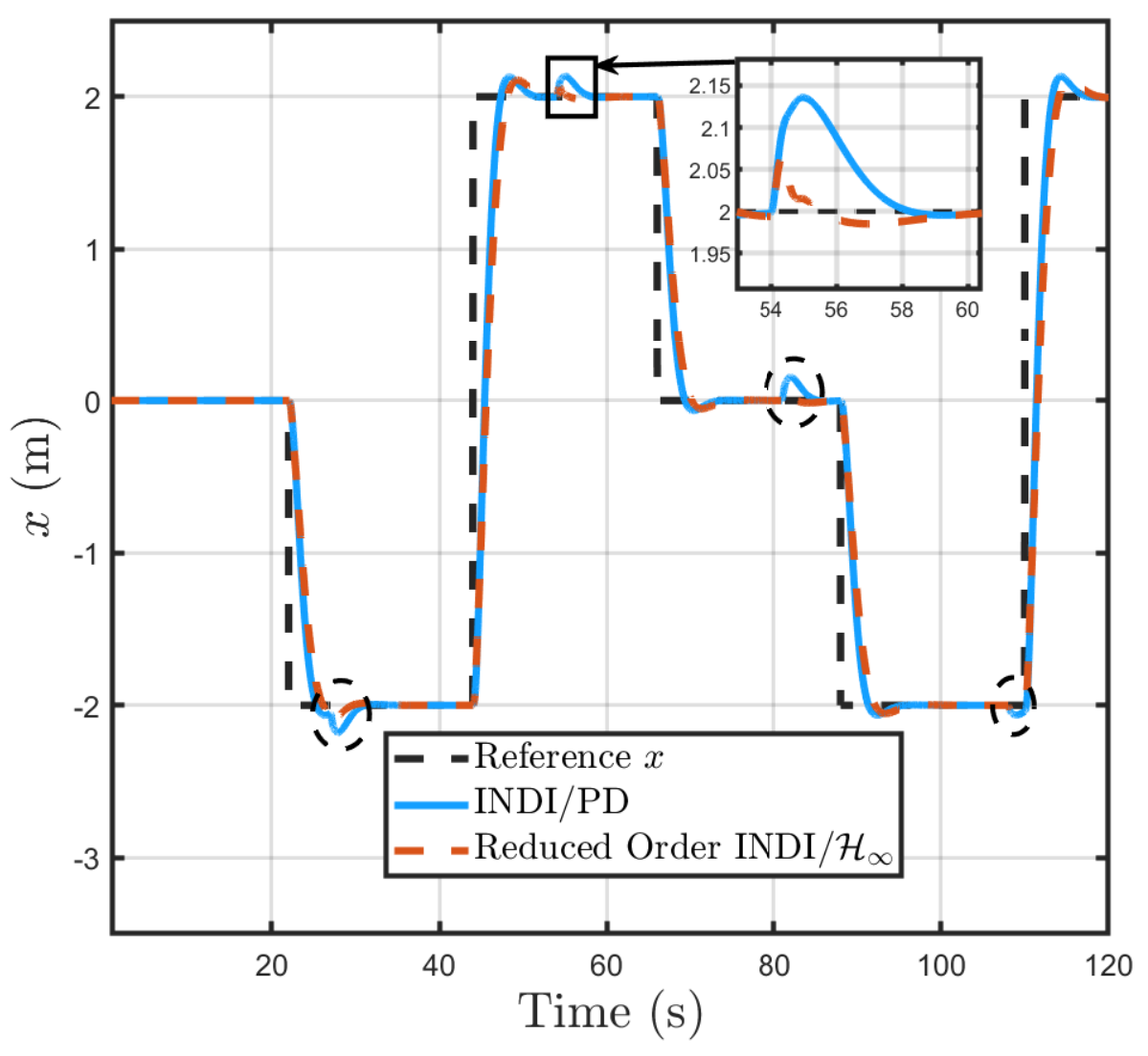

(a) Position tracking

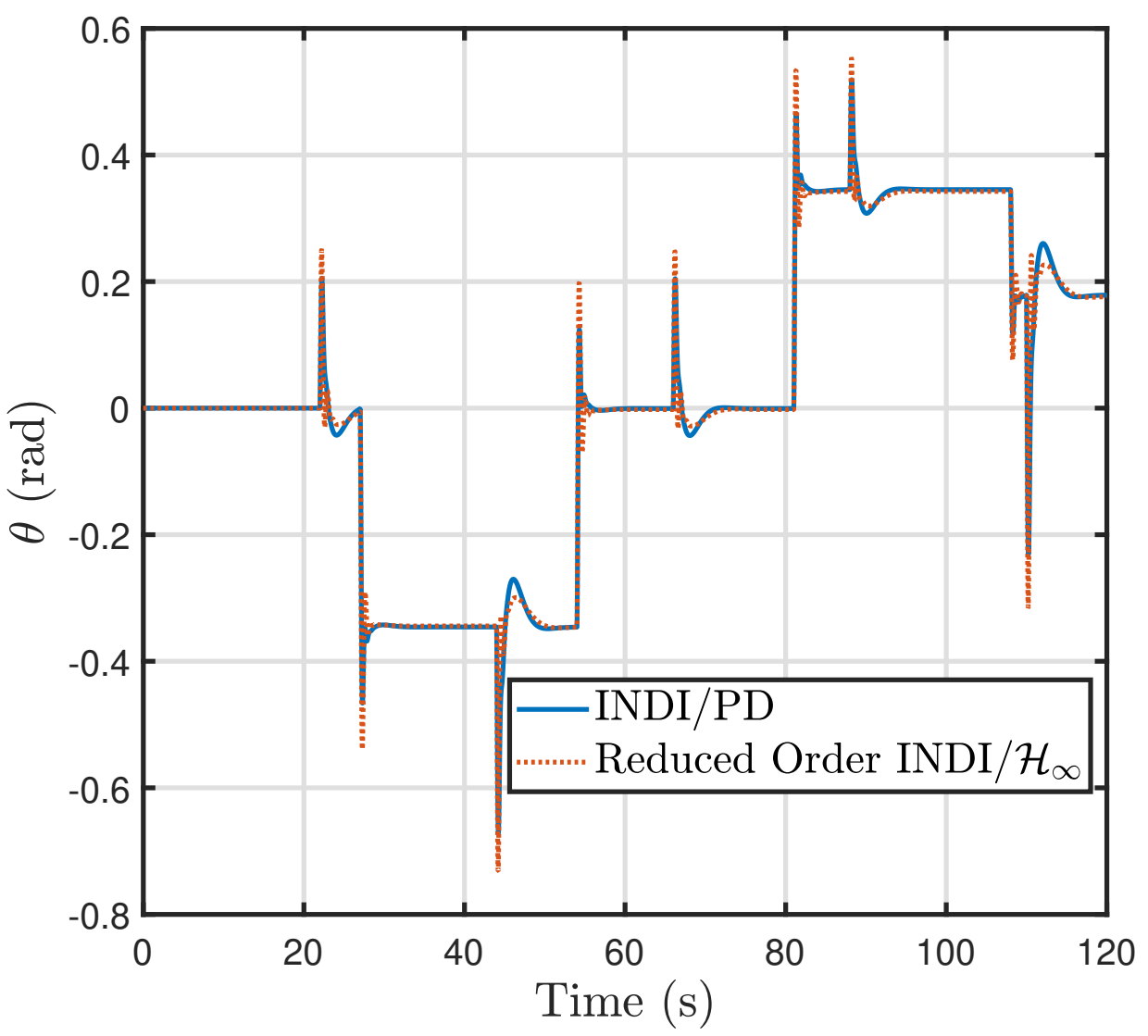

(b) Pitch angle

Fig. 10: Position tracking in the presence of disturbances highlighted by the rectangle and the dashed ellipses.

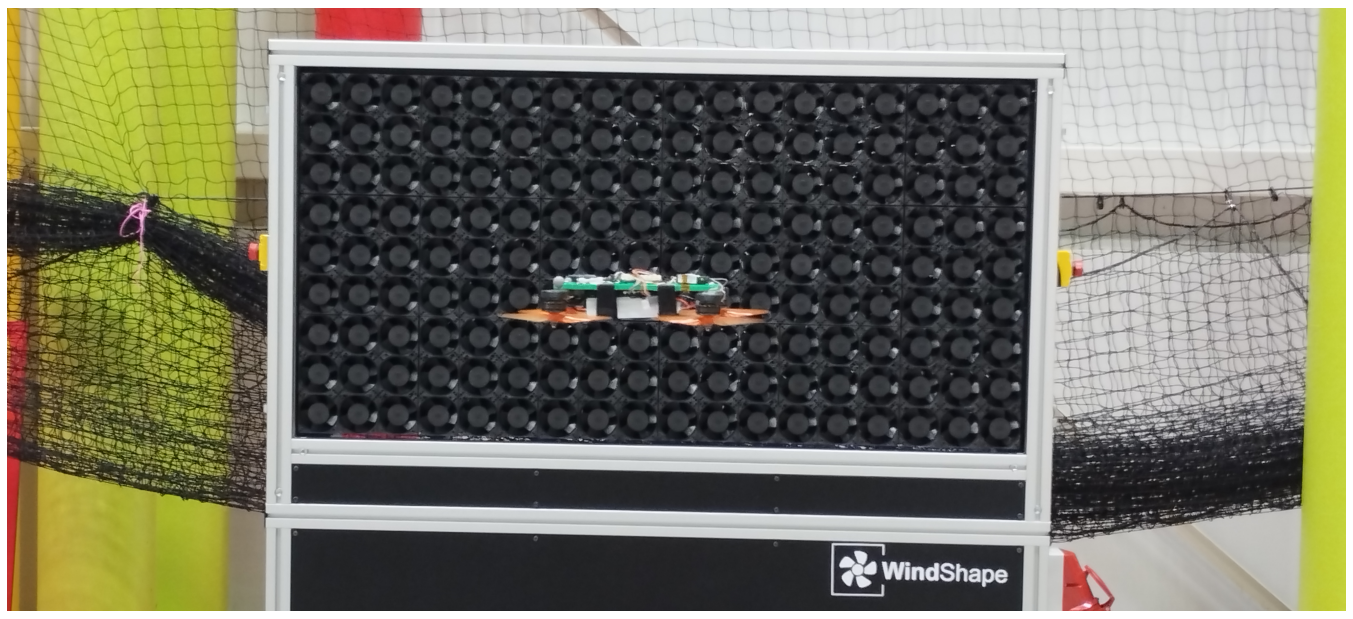

Fig. 11: Quadcopter hovering in front of the wind generator.

The controllers of the guidance and the stabilization loops are then designed as explained in Sections IV-A and IV-B respectively. The values of the weighting templates used in the experiments differ from those used in the simulations due to the differences in the used actuators, as shown in Table III. The estimated actuator model has a time constant $\hat{\tau}_m$ almost twice larger than the one used in the simulations ($\tau_m = \frac{1}{53.94}$ s), resulting in different tracking speeds. All controllers are designed in the continuous time domain and then discretized using the Tustin method at a frequency of 500 Hz.

To have a fair comparison between the designed $\mathcal{H}_\infty$ controller and a well-tuned PD controller for both loops, a PD controller is first tuned for the inner loop using modal control, based on the estimated model $\hat{A}(s)$. The closed-loop poles are placed to have a damping ratio of 0.6 and a natural frequency of 14 rad/s, which results in P and D gains equal to $K_\mu^{PD} = 5.2$ and $K_\Omega^{PD} = 13.3$. The closed-loop poles of the inner loop are then placed at $-8.85$ rad/s and $-8.4 \pm 11.2i$ rad/s. It is worth mentioning that we have tuned this PD controller using the proposed $\mathcal{H}_\infty$ design framework, forcing $K_\mu$ and $K_\Omega$ to be static gains. We selected identical weighting templates ($W_e(s)$ and $W_u$) as those used in the design of the second order $\mathcal{H}_\infty$ controller, defining same performances (tracking error, avoiding saturation), while excluding the disturbance and noise performance terms to obtain a feasible solution. It turns out that the computed control gains and the closed-loop poles are almost exactly the same as those obtained with modal control, which confirms that this choice of PD controller is relevant. On the other side, the PD controller of the outer loop is tuned considering the GPS update frequency and the speed of the inner closed-loop. The goal is to achieve the same bandwidth than the second order $\mathcal{H}_\infty$ controller, with a minimum overshoot. The PD gains are then chosen as $K_\xi^{PD} = 0.8$ and $K_v^{PD} = 1.8$.

TABLE III: Weighting templates used in the experiments

| Weighting Templates | Stabilization Loop | Guidance Loop |
|---|---|---|
| $W_e(s)$ | $\frac{\frac{s}{2} + 2\pi}{s + 2\pi \times 0.0001}$ | $\frac{\frac{s}{2} + 2\pi 0.15}{s + 2\pi 0.15 \times 0.0001}$ |
| $W_u$ | 400 | 2 |
| $W_d(s)$ | $10\, W_e(s)$ | $2\, W_e(s)$ |
| $W_n$ | 0.1 | 0.1 |

## B. Hovering in the Presence of Gusts

The first scenario consists of hovering in front of the wind generator. The wind speed varies, starting at 3.6 m/s and then gradually increasing every 20 s to 7.2 m/s and 10.8 m/s. This aims to provide a persistent perturbation and validates the drone's ability to attenuate perturbations at different speeds. To check the ability of the drone to reject sudden disturbances, we introduced a disturbance of gust with a speed of around 9 m/s. The position of the drone while hovering in front of the wind generator employing INDI/$\mathcal{H}_\infty$ controller is shown in Fig. 12a. The effect of each exerted disturbance on the drone's position is indicated by the dotted ellipses. It is shown that for a persistent perturbation, the drone's lateral deviation with the INDI/$\mathcal{H}_\infty$ controller is about 0.15 m, while that for the last sudden disturbance is about 0.3 m within the $x$ axis. On the other hand, Fig.12b shows higher deviation in the lateral $x$ axis of the drone while hovering with the INDI/PD controller, reaching to 0.45 m for the first perturbation. The drone was able to reject a sudden disturbance of wind speed equal to 9 m/s but with a lateral deviation in the $x$ direction up to 1 m. The wind should ideally only affect the drone along the $x$ axis, but turbulence and indoor experimental conditions also cause some perturbations along the $y$ axis. The INDI/$\mathcal{H}_\infty$ controller significantly reduces these effects, while the INDI/PD controller shows a 0.15 m deviation on the $y$ axis. It is also observed that tracking in the $z$ axis for the INDI/$\mathcal{H}_\infty$ controller is better than the tuned INDI/PD controller, as shown in Figs. 12a and 12b. This can be attributed to the fact that the estimated thrust $f_B$ is calculated depending on the estimated $\hat{G}_{12}$ which is not accurate and includes some uncertainties. Video of the experiment can be seen by referring to the provided link [3].

[3]https://drive.google.com/file/d/1O4jDks1QgUeJbdtbgitgAsBIghxwnxdt/view?usp=sharing

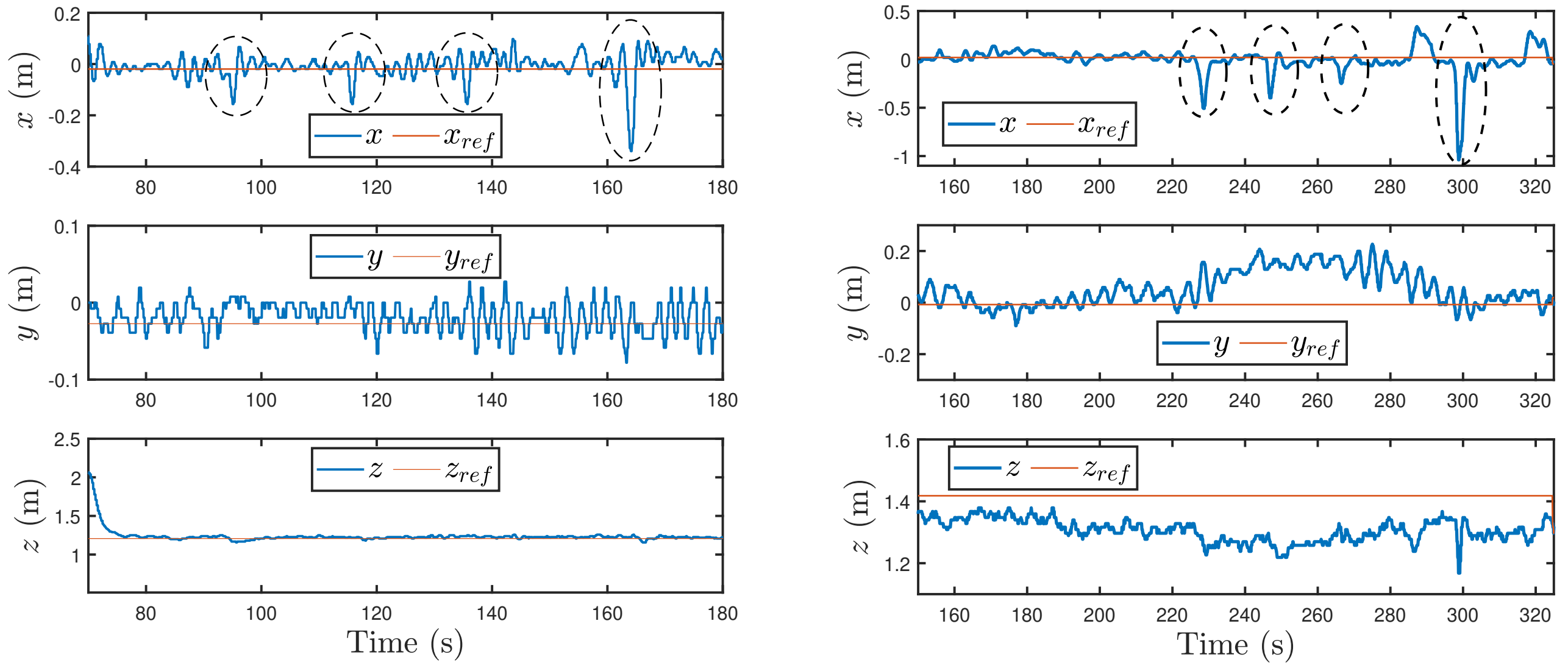

(a) INDI/$\mathcal{H}_\infty$

(b) INDI/PD

Fig. 12: Drone hovering in front of the wind generator. The effects of disturbances are surrounded by ellipses.

### *C. Passing through Defined Waypoints with Gusts*

The second scenario involves the drone navigating through three defined waypoints in a triangular geometry. The gust is created by the wind generator in the drone's path, leading it through areas where there is a constant wind speed of 7.2 m/s. The objective is to observe the behavior of the drone while flying into a suddenly turbulent field and being affected by sudden disturbances during the flight time. First flight is done in the absence of disturbances, and then four consecutive flights are done between the three defined waypoints. It is shown in Fig. 13a that while using the INDI/$\mathcal{H}_\infty$ controller, there is a deviation reaching up to 0.4 m, especially when the drone is near the wind generator. However, it is noted that the deviation is smaller when the drone re-enters the disturbance zone while flying diagonally. This is primarily because it encounters slower wind compared to the first entry, and flying diagonally distributes the disturbances across both the $x$ and $y$ axes of the drone. The same test is conducted to validate the tuned INDI/PD controller. The deviation due to the wind disturbance in this case reaches up to 0.75 m while passing just in front of the wind generator. It is also noted from Fig. 13b that the deviation of the drone during the diagonal leg in front of the wind generator is also high, reaching up to 0.9 m. Video of the experiment can be seen by referring to the provided link[4].

## VI. Conclusion

A new methodology is proposed in this work to design a robust INDI/$\mathcal{H}_\infty$ architecture for aerial robotics applications, which maximizes the system's robustness against external disturbances such as wind or gust. The design of the $\mathcal{H}_\infty$ controllers assumes that these disturbances affect the drone's inputs, therefore the linear controller gains play a major role in their attenuation. A cascaded architecture is employed for both the stabilization and guidance controllers to manage the system's dynamics. A structured $\mathcal{H}_\infty$ control problem is formulated and solved using non-smooth optimization techniques to obtain low-order controllers, which are scarcely more complex than the PD controllers classically proposed in the literature. A comparative analysis is first conducted between the proposed INDI/$\mathcal{H}_\infty$ and the existing INDI/PD approaches via simulations using a Parrot Bebop quadrotor model. The proposed architecture significantly improves the system's ability to reject disturbances, by more than 50% for both the rotational and translational dynamics. Then, all results are validated experimentally in front of a wind generator using a quadcopter drone built by the ENAC team. Overall, the proposed method improves the system's robustness to external disturbances and allows to handle significant modeling uncertainties in the estimated control effectiveness matrices and actuator model. Future work will involve the use of $\mu$-analysis to better analyze the effects of these uncertainties, and provide insights to improve the tuning of the $\mathcal{H}_\infty$ controllers, thus further improving the robustness of the system.

[4]https://drive.google.com/file/d/1TB8xuwno7Fo016U4Ulym6_Ki_siIivpd/view?usp=sharing

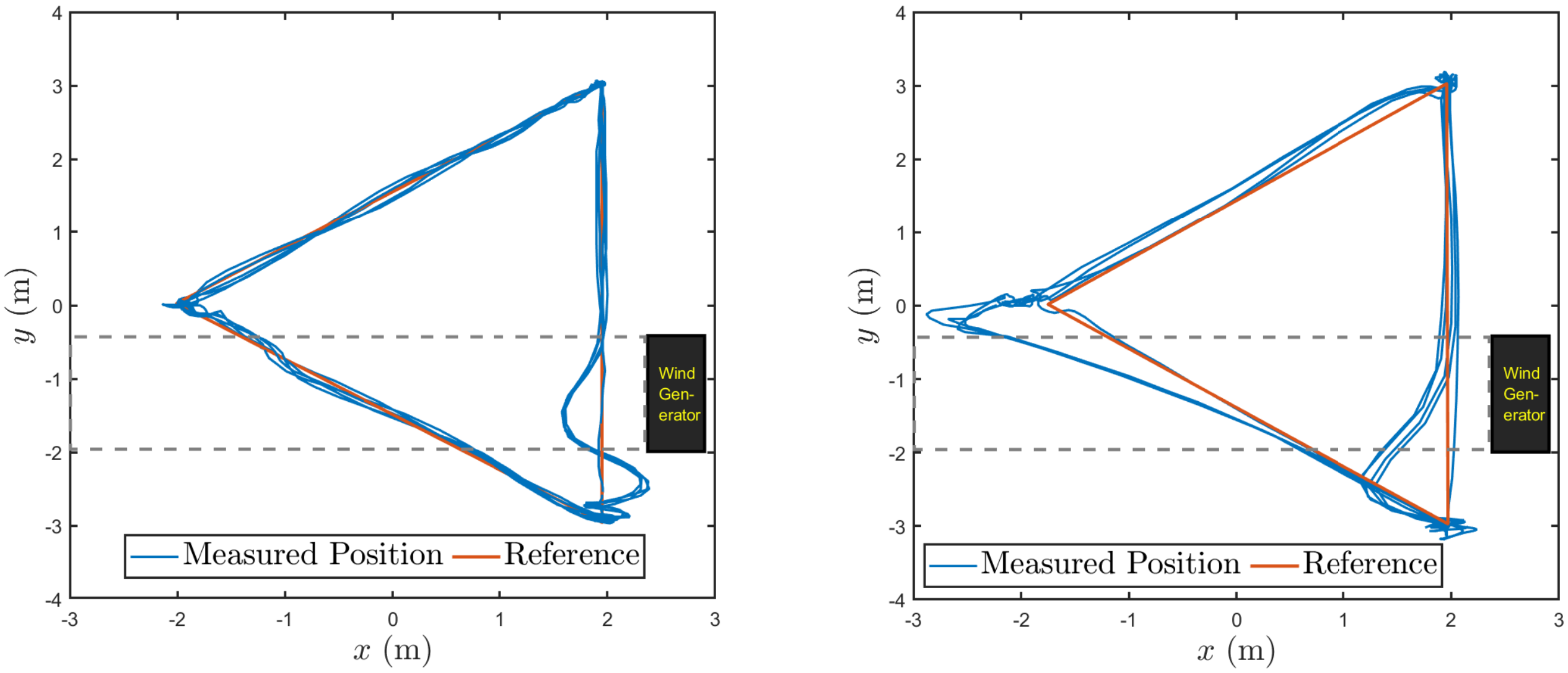

(a) INDI/$\mathcal{H}_\infty$ (b) INDI/PD

Fig. 13: Position tracking of the drone while passing through defined waypoints in the presence of wind.